\documentclass[letterpaper, 10 pt, conference]{ieeeconf}  % Comment this line out if you need a4paper

\IEEEoverridecommandlockouts                              % This command is only needed if 
\title{\LARGE \bf
Model Predictive Loitering and Trajectory Tracking of Suspended Payloads in Cable-Driven Balloons Using UGVs     
}

\author{Julius Wanner$^{1}$, Eric Sihite$^{2}$, Alireza Ramezani$^{3}$, Morteza Gharib$^{2}$% <-this % stops a space
\thanks{$^{1}$ The author is a visiting researcher at the Graduate Aerospace Laboratories of the California Institute of Technology (Caltech GALCIT), Pasadena, CA-91125, USA and with the Department of Mechanical and Process Engineering, ETH Zurich, CH-8092 Zürich, Switzerland.
        {Email: \tt\small jwanner@student.ethz.ch}}%
\thanks{$^{2}$The authors are with the Graduate Aerospace Laboratories of the California Institute of Technology (Caltech GALCIT), Pasadena, CA-91125, USA.
        {Emails: \tt\small esihite, mgharib@caltech.edu}}%
\thanks{$^{3}$ The author is with the SiliconSynapse Laboratory, Department of Electrical and Computer Engineering, Northeastern University, Boston, MA-02119,USA.
    {Email: \tt\small a.ramezani@northeastern.edu}}%
}

\usepackage{amsmath}
\usepackage{graphics}
\usepackage{graphicx}
\usepackage{amssymb}
\usepackage{color}
\usepackage{upgreek}
\usepackage{bbm, bbold}
\usepackage{algorithm}
\usepackage{algpseudocode}
\usepackage{lipsum}
\begin{document}

\maketitle
\thispagestyle{empty}
\pagestyle{empty}

%%%%%%%%%%%%%%%%%%%%%%%%%%%%%%%%%%%%%%%%%%%%%%%%%%%%%%%%%%%%%%%%%%%%%%%%%%%%%%%%
\begin{abstract}
The feasibility of performing airborne and ground manipulation, perception, and reconnaissance using wheeled rovers, unmanned aerial vehicles, CubeSats, SmallSats and more have been evaluated before. Among all of these solutions, balloon-based systems possess merits that make them extremely attractive, e.g., a simple operation mechanism and endured operation time. However, there are many hurdles to overcome to achieve robust loitering performance in balloon-based applications. We attempt to identify design and control challenges, and propose a novel robotic platform that allows for the application of balloons in the reconnaissance and perception of Mars craters. This work briefly covers our suggested actuation and Model Predictive Control design framework for steering such balloon systems. We propose the coordinated servoing of multiple unmanned ground vehicles (UGVs) to regulate tension forces in a cable-driven balloon to which an underactuated hanging payload is attached.
\end{abstract}

%%%%%%%%%%%%%%%%%%%%%%%%%%%%%%%%%%%%%%%%%%%%%%%%%%%%%%%%%%%%%%%%%%%%%%%%%%%%%%%%
\section{INTRODUCTION}
Building on a background of successful high-altitude balloon usage in atmospheres for  reconnaissance, climate monitoring , and planetary exploration \cite{sagdeev_vega_1986} to name a few, in this work we aim at identifying design and control challenges, and propose a novel robotic platform that renders the increased use of balloons in space possible, particularly for suggested reconnaissance and perception applications around Martian craters \cite{meirion-griffith_accessing_2018}. In this way, we initiated a collaborative project between Caltech's Center for Autonomous Systems and Technologies (CAST) and Northeastern University's \textit{SiliconSynapse} Lab that inspects the full fledged design, modeling and control of an autonomous robot for performing tasks such as perception, manipulation, or construction in martian craters. This work briefly covers our proposed actuation and control design framework based on the coordinated servoing of multiple unmanned ground vehicles (UGVs) to regulate tension forces in a cable-driven balloon with an underactuated tethered payload.

The feasibility of performing communication using Satelite Signals of Opportunity (SoOp) \cite{lally_tethered_2020}, airborne and ground manipulation and reconnaissance with unmanned aerial vehicles (UAVs), UGVs, high altitude balloons (HAB), SmallSats (SS), and CubeSats (CS) have been evaluated before \cite{meirion-griffith_accessing_2018,gamber_mars_1997,anyoji_evaluation_2017,takemura_development_2005,poghosyan_cubesat_2017},  and first aerial systems have recently been flight-proven on Mars \cite{balaram_ingenuity_2021}.  Among all of these solutions, balloon-based systems possess merits that make them extremely attractive. A balloon's simple operation mechanism and endured operation time with minimum energy throughput has made them well-suited for space applications. However, there are many hurdles to overcome to achieve robust loitering performance in balloon-based applications.

\begin{figure}[t]
\centering
\vspace{0.1in}
    \includegraphics[width=\columnwidth]{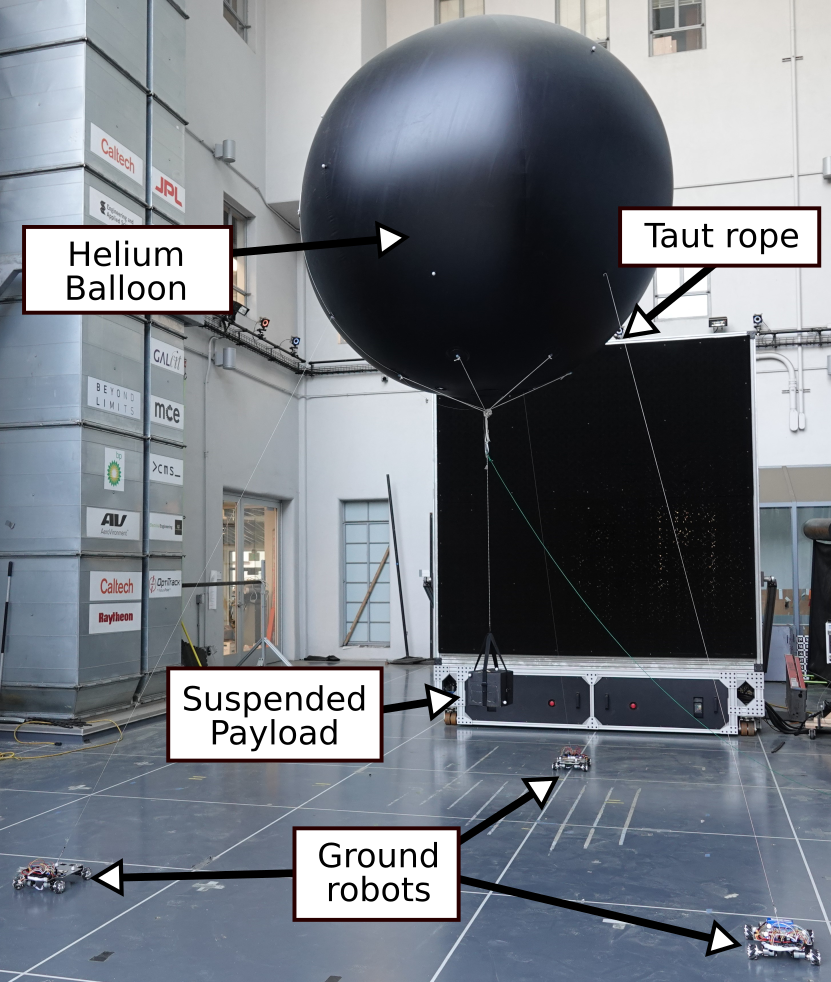}
    \caption{Illustrates our cable-driven balloon prototype with underactuated hanging payload controlled based on the coordinated servoing of multiple UGVs to regulate tether tension forces.}
    \label{fig:cover-image}
\vspace{-0.1in}
\end{figure}

The environment has the first-order impact on balloon design and control. There are several obvious reasons to strongly consider environmental effects. Briefly speaking, these atmospheric effects include (1) circulation dictating balloon ground track, (2) density determining balloon size, and (3) radiation affecting balloon envelope strength, and (4) dust, thermal inertia, and surface albedo indirectly introduced by the Martian surface \cite{garg_balloon_2020}. These challenges collectively make steering these balloons very challenging. For instance, the Martian environment is famous for its dust storms partly because the Martian surface is covered with porous regolith. Dusty atmosphere moderates the effects of solar and surface radiation by increasing air opacity \cite{gamber_mars_1997}. With moderate radiation, temperature fluctuations which are key to drastic changes in balloon dynamics are minimized. However, on Mars, dust can pose challenges to any fan-based balloon control methods that have been explored so far on Earth, where the environment is far less hostile.

In addition, although dynamic models and altitude control methods for planetary atmospheric balloons have been suggested previously \cite{garg_balloon_2020,hall_prototype_2021,schuler_altitude_2020}, changing atmospheric conditions and highly nonlinear balloon-atmosphere interactions in form of friction and pressure drag forces make precise trajectory control of these floating structures extremely challenging \cite{renegar_survey_2017}. Any balloons that are operated at low altitudes close to the Martian surface will furthermore see the full diurnal variation of Martian temperature. If the explored area is at higher altitudes, such as the Garni Crater with an elevation of $\sim$ 5000 m \cite{meirion-griffith_accessing_2018}, and ground patches with low thermal inertia are traversed, the variations in balloon gas temperature are further accentuated \cite{gamber_mars_1997} and thus affect internal pressure, volume and net lift. Therefore, nonlinear control design paradigms that could predict plant behavior under reasonable dynamical models can mitigate model uncertainty and make model-based nonlinear control of these robots feasible. Note that model-based nonlinear control design paradigms have a rich history in engineering problems and in the past used to assume the existence of perfect mathematical models. The application of model-based control \cite{dangol_towards_2020,dangol_thruster-assisted_2020,ramezani_modeling_2016,buss_preliminary_2014,ramezani_towards_2020,ramezani_biomimetic_2017,ramezani_describing_2017,ramezani_atrias_2012} to slow and fast dynamical robots in recent years with the advent powerful embedded computers has sharply increased and now opens new horizons for other applications such as balloon control. Based on our past experience, a first model-based control strategy is designed for constant lift helium-filled balloons in this work in order to investigate trajectory tracking feasibility,

\section{System Overview}

The proposed prototype is depicted in Fig.~\ref{fig:cover-image}. The helium-filled balloon constructed from polyvinyl chloride has an empty structural mass of $\sim$ 3.6 kg and a diameter of 2.2 m, such that it is capable to lift a payload of up to 2 kg. The illustrated payload has a mass of 1 kg throughout testing and is attached vertically beneath the balloon's center. The shown two-part tethering configuration is used in order to reduce unwanted pitch and roll movements of the payload about its tether fixation point.

Furthermore, three mecanum-wheel omnidirectional UGVs built from an aluminum frame are tethered to the balloon via a thin string. The robot-balloon tethers are directly fixed to the balloon surface at an elevation angle of 30° from the equator with equal azimuth angles of 120° between each other. Every ground robot possesses a WiFi module that receives control commands in form of serial communication packets from a central control machine and transmits these to an onboard microcontroller. The microcontroller then generates low-level signals that drive the UGV omnidirectional wheels. The central controller is run on a PC and utilizes \textit{Simulink Desktop Real-Time (SDRT)} software to generate and send control inputs. The use of SDRT permits seamless implementation of simulated control algorithms on robot hardware.

\section{Dynamic model}

In this section, we derive the nonlinear equations of motion for the proposed system. To allow for simulation and control design using MATLAB and Simulink, a three-dimensional continuous-time model is developed. 
% please don't delete this gap!

\begin{figure}[t]
\begin{center}
\vspace{0.1in}
    \includegraphics[width=\columnwidth]{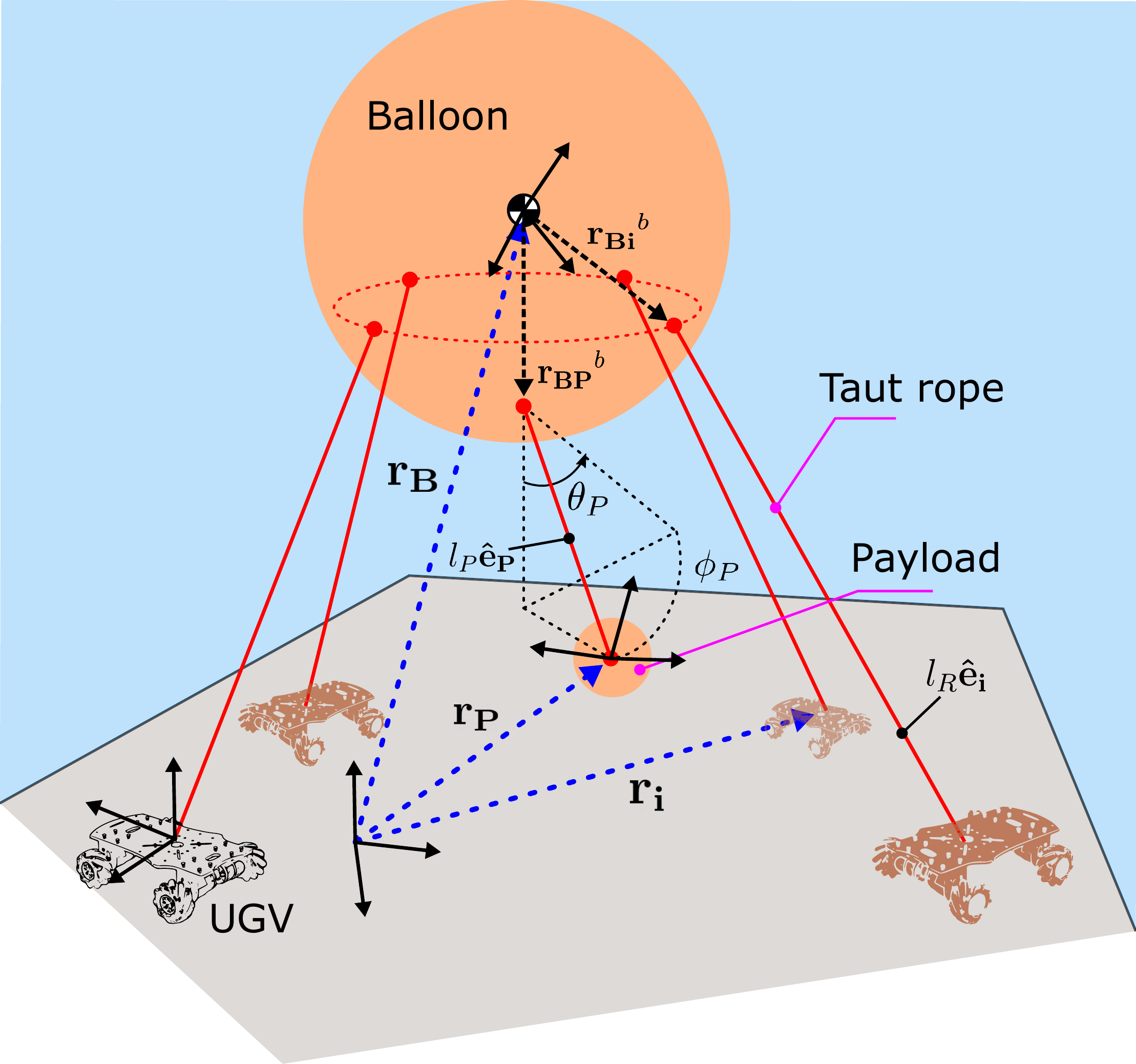}
    \caption{Model diagram of balloon, payload, and mecanum-wheel UGVs.}
    \label{fig:FBD_1}
\vspace{-0.1in}
\end{center}
\end{figure}

The balloon is modeled as a 6-DOF rigid body. The vectors $\mathbf{r_{B}} = [x_{B}, y_{B}, z_{B}]^\top$ and $\mathbf{\Theta_{B}} = [\phi_{B}, \theta_{B}, \psi_{B}]^\top$ are chosen to represent the inertial position of the balloon's center of mass as shown in Fig.~\ref{fig:FBD_1} and its attitude Euler angles, respectively. The tethered payload is assumed as a point-mass single pendulum, where its motion relative to the balloon is described with two coordinates $\mathbf{\Theta_{P}} = [\phi_{P}, \theta_{P}]^\top$ and the payload position in the inertial reference frame is denoted by $\mathbf{r_{P}} = [x_{P}, y_{P}, z_{P}]^\top$. These coordinates are illustrated in Fig.~\ref{fig:FBD_1}.
% please don't delete this gap!

The omni-directional UGVs are constrained to move on the ground plane and thus each posess three controllable degrees of freedom. To reduce model complexity for controller development, each UGV motion is analyzed with a configuration kinematic model (CKM). The CKMs offer a mathematical model that relates the kinematic states of the UGVs to the physically commanded inputs.The UGV center of mass positions $\mathbf{r_{i}} = [x_{i}, y_{i}, z_{i}]^\top, ~i \in \left\{ 1,\dots, n\right\}$ (where $n$ denotes the total number of ground rovers) as portrayed in Fig.~\ref{fig:FBD_1} and their heading angle about the z axis, denoted by $\theta_{i}$, are used for developing each CKM. 

As only one tether attaches the balloon to the pendulum, the payload is underactuated. The payload and robot tethers have respective constant lengths denoted by $l_{P}$ and $l_{R}$ and their respective directions are parameterized by unit vectors $\mathbf{\hat{e}_{P}}$ and $\mathbf{\hat{e}_{i}}$. The quantities $\mathbf{r_{BP}}^{b}$ and $\mathbf{r_{Bi}}^{b}$ illustrated in Fig.~\ref{fig:FBD_1} describe the distance vectors starting from the balloon center of mass and directed towards the payload attachment point and the i-th UGV tether attachment point respectively.

% placeholder used for fast complilation

The Newton-Euler approach is employed to model only the balloon and tethered payload subsystem.
For a balloon of structural mass $m_{B}$, volume $V_{B}$ and body-fixed moment of inertia tensor $\mathbf{I_{B}}$, the translational and rotational dynamics in the inertial and body frames are obtained. The upwards directed net balloon buoyancy force is defined as
\begin{equation}
F_{B,z} = (\rho_{a} - \rho_{g})V_{B}g - m_{B}g
\label{eq:buoy_eq}
\end{equation}
where $g$, $\rho_{a}$ and $\rho_{g}$ denote the gravitational acceleration, atmospheric density and density of the gas contained in the balloon, respectively. The remaining external forces acting on the balloon are the aerodynamic drag force, $\mathbf{F_{D,B}}$, the pendulum (payload) tether tension force, $\mathbf{T_{p}}$, and the tether tension force between the balloon and the i-th UGV, $\mathbf{T_{i}}$. 

The unilateral tether forces can only act in the direction of the tethers and are thus parameterized based on their magnitude and direction, as shown in \eqref{eq:bal-dyn}. Note that while complex unilateral tether models could be used in our study, we limit ourselves to tether connections that are inelastic and massless strings which can only support tensile forces. 

The overall dynamics of the balloon are thus given by
\begin{equation}
\begin{aligned}
% \begin{bmatrix}
%     m_{B}\ddot{x}_{B} \\
%     m_{B}\ddot{y}_{B}  \\
%     (m_{B}+m_{v})\ddot{z}_{B} \\ 
% \end{bmatrix} =
& m'_{B}\mathbf{\ddot r_B}=
\mathbf{F_{D,B}}(\mathbf{\dot{r}_{B}}) + \mathbf{T_{P}} + \sum_{i=1}^{n} \mathbf{T_{i}} \ + \mathbf{W'}\\
% \begin{bmatrix}
%     0 \\
%     0 \\
%     F_{B,z} - m_{B}g \\
% \end{bmatrix} \\
& \mathbf{I_{B}}\mathbf{\dot{\omega}_{B}}^{b} + \mathbf{{\omega}_{B}}^{b} \times  \mathbf{I_{B}}\mathbf{{\omega}_{B}}^{b} = \mathbf{M_{P}}^{b} + \sum_{i=1}^{n} \mathbf{M_{i}}^{b} \,\\
& \mathbf{T_{P}} =  \|\mathbf{T_{P}}\|\mathbf{\hat{e}_{P}} , \; \mathbf{T_{i}} = \|\mathbf{T_{i}}\|\mathbf{\hat{e}_{i}} \\
& \mathbf{M_{P}}^{b} = (\mathbf{r_{BP}}^{b} \times  \mathbf{R_{0b}}\mathbf{T_{P}}) \\    & \mathbf{M_{i}}^{b} = (\mathbf{r_{Bi}}^{b} \times  \mathbf{R_{0b}}\mathbf{T_{i}})
\end{aligned}
\label{eq:bal-dyn}
\end{equation}
\newline
\noindent where $\mathbf{W'} = [0, 0, F_{B,z}]^\top$ and an added mass arising from the air mass that is moved along with the balloon \cite{garg_balloon_2020} is included in the term $m'_{B}$. The symbols $(.)^b$, $\|.\|$ and $\times$ denote a quantity in balloon body-fixed frame, the Euclidean norm, and the vector cross-product respectively. The angular velocity $\mathbf{\omega}_{B}^{b}$ notably is used to define the roll, pitch and yaw rates in body-fixed frame. The homogeneous transformation matrix $\mathbf{R_{0b}}$ used in \eqref{eq:bal-dyn} transforms the tether forces and moments from the inertial system into the balloon body frame. 

To identify the dynamics of the payload, in the following procedure we derive explicit equations for the angular accelerations $\mathbf{\ddot{\Theta}_{P}} = [\ddot{\phi}_{P}, \ddot{\theta}_{P}]$ and the tether tension force magnitude $\|\mathbf{T_{P}}\|$. First, in \eqref{eq:pay_pos_form} the payload position in the inertial frame ($\mathbf{r_{P}}$) is expressed using geometric relations known from the balloon model. Here, the unit vector $\mathbf{\hat{e}_{P}}$ is expressed solely as a function of the payload coordinates $\mathbf{\Theta_{P}}$, whereas all other quantities are dependant only on the balloon coordinates  $\mathbf{r_{B}}$ and $\mathbf{\Theta_{B}}$. 
The first and second order derivatives of \eqref{eq:pay_pos_form} are consequently formed to yield expressions for $\mathbf{\dot{r}_{p}}$ and $\mathbf{\ddot{r}_{p}}$. The three-dimensional equations of motion of the payload in the inertial system are formulated and presented in \eqref{eq:pay-dyn} , where $\mathbf{F_{D,P}}$ represents the aerodynamic drag acting on the payload alone and $\mathbf{W_{P}}$ is the pendulum weight vector.
\begin{align}
\mathbf{r_{P}} = \mathbf{r_{B}} + (\mathbf{R_{0b}})^T\mathbf{r_{BP}}^{b} + l_{P}\mathbf{\hat{e}_{P}} \label{eq:pay_pos_form} \\
m_{P}\mathbf{\ddot{r}_{P}} = \mathbf{F_{D,P}}(\mathbf{\dot{r}_{P}}) - \mathbf{T_{P}} - \mathbf{W_P} \label{eq:pay-dyn}
\end{align}
\noindent Next, the introduced expression for $\mathbf{T_{P}}$ in \eqref{eq:bal-dyn} and the quantities $\mathbf{r_{P}}$,  $\mathbf{\dot{r}_{p}}$ and $\mathbf{\ddot{r}_{p}}$ obtained in previous steps are substituted into \eqref{eq:pay-dyn}. After further substitution of the quantities $\mathbf{\ddot{\Theta}_{B}}$ and $\mathbf{\ddot{r}_{B}}$ solved in \eqref{eq:bal-dyn}, the result is rearranged for the three payload variables $\ddot{\phi}_{P}, \ddot{\theta}_{P}, \|\mathbf{T_{P}}\|$.  

The combined balloon-payload dynamics are obtained and are given by
\begin{equation}
\begin{aligned}
    \mathbf{x_{BP}}(t) &= [\, \mathbf{r_{B}}^\top,\mathbf{\Theta_{B}}^\top, \mathbf{\Theta_{P}}^\top,\mathbf{\dot{r}_{B}}^\top,\mathbf{\dot{\Theta}_{B}}^\top, \mathbf{\dot{\Theta}_{P}}^\top \, ]^\top \\
    \mathbf{\dot{x}_{BP}}(t) &= f(\mathbf{x_{BP}}(t),\mathbf{T_{i}}) \\
\end{aligned}
\label{eq:DAE}
\end{equation}
where the nonlinear function $f(\mathbf{x_{BP}}(t),\mathbf{T_{i}})$ is dependant on the state vector $\mathbf{x_{BP}(t)}$ and steered by the unilateral tether forces between the balloon and the UGVs, i.e., $\mathbf{T_{i}}$. 

\subsection{UGV Configuration Kinematics Model (CKM)}

The selected UGVs use mecanum wheels (also referred to as Swedish wheels) to facilitate omni-directional motion. As the robots are modeled using a CKM formulation, the inputs for the purpose of higher level control are chosen to be the second time derivative of planar position and heading angles of the UGVs.

The control inputs of the i-th UGV are thus denoted by $\mathbf{u_{i}} = [{u}_{i_1}, {u}_{i_2}, {u}_{i_3}]^\top$, where $u_{i,1}$,$u_{i,2}$ and $u_{i,3}$ command the respective rover accelerations $\ddot{x}_{i}$, $\ddot{y}_{i}$ and $\ddot{\theta}_{i}$ in the inertial reference frame. 

Nonetheless, though this input structure can be implemented in simulation, in the physical system the input accelerations $\mathbf{u_{i}}$ cannot be directly actuated. The four mecanum-wheel rotational speeds of every UGV can however be independently controlled on a lower-level such that the commanded accelerations are achieved. The CKM is hence formulated to not only derive the kinematic states of the UGVs in the inertial frame, but to further obtain a set of four wheel angular velocity references for each UGV. These can be derived from the forward kinematics of mecanum-wheel rovers obtained in other work \cite{taheri_hamid_kinematic_2015}.
The CKM formulation relating the acceleration inputs $\mathbf{u_{i}}$ to the kinematic states and the wheel angular velocities of the i-th UGV is in summary given by:
\begin{equation}
%\left\{
\begin{aligned}
\mathbf{x_{R,i}}(t) &= [x_{i}, y_{i}, \theta_{i}, \dot{x}_{i}, \dot{y}_{i}, \dot{\theta}_{i}]^\top \\
\mathbf{\dot x_{R,i}}(t) &= [\dot{x}_{i}, \dot{y}_{i}, \dot{\theta}_{i}, {u}_{i_1}, {u}_{i_2}, {u}_{i_3}]^\top \\
% \mathbf{v^b_{i}} &= \begin{bmatrix}\cos{\theta_{i}} & \sin{\theta_{i}} \\ -\sin{\theta_{i}} & \cos{\theta_{i}} \end{bmatrix}
% \begin{bmatrix} \dot{x}_{i} \\ \dot{y}_{i} \end{bmatrix} \\
\mathbf{v^R_{i}} &= \mathbf{R_{0i}}(\theta_i)
[\dot{x}_{i}, \dot{y}_{i}]^\top \\
\mathbf{\omega_{w}^{i}} &= \mathbf{T_{kin}}[\mathbf{v^R_{i}}^\top, \dot{\theta}_{i}]^\top
\label{eq:CKM}
\end{aligned}
%\right.
\end{equation}
\noindent where $\mathbf{x_{R,i}}(t)$ denotes the kinematic states used for simulation in the inertial frame, the column vector $\mathbf{\omega_{w}^{i}}$ denotes the four mecanum wheel angular velocities of the i-th UGV and $\mathbf{v^R_{i}}$ is the planar velocity vector in the i-th UGV body frame. The matrices $\mathbf{R_{0i}}$ and $\mathbf{T_{kin}} \in \mathbb{R}^{4\times`3}$ respectively denote a 2D planar rotation matrix transforming into body frame and the forward kinematics transformation \cite{taheri_hamid_kinematic_2015}. In \eqref{eq:CKM}, we make the assumption that the center of gravity coincides with the robot's geometric center. 

%In Eq.~\ref{eq:CKM}, it is assumed that the tethers are attached to the center of gravity of the robot and that the center of gravity coincides with the robot's geometric center. Variations to this assumption can be made if a large priority is given to actuate the robot rotational acceleration as a means to control the system.

\subsection{Unilateral Tether Force Constraints}

In reality, in addition to being underactuated, the unilateral tether tension forces yield switching dynamics. We overlook the switching effects by enforcing a set of constraints when resolving control actions. The magnitudes of the balloon-UGV tension forces $\|\mathbf{T_{i}}\|$ are obtainable from forward kinematics by assuming the tethers always remain taut, hence through enforcing that each distance vector between the two tether ends, denoted by $\mathbf{\Delta r_i}$, has a magnitude that is always constant and equal to the tether length. This purely state dependant distance constraint is expressed with a function $K_{i}$. After double-differentiation and further state variable substitutions a set of $n$ algebraic equations (constraints) denoted by the functions $k_{i}, ~i \in \left\{ 1,\dots, n\right\}$, which are implemented in simulation, is obtained. The tether geometric quantities and constraint function expressions are summarized in \eqref{eq:taut-tether-constraint} below
\begin{equation}
\begin{aligned}
&\mathbf{\Delta r_i}(\mathbf{x_{BP}}(t),\mathbf{x_{R,i}}(t)) = \mathbf{r_{i}} - (\mathbf{r_{B}} + (\mathbf{R_{0b}})^T\mathbf{r_{Bi}}^{b}) \\
    &K_{i}(\mathbf{x_{BP}}(t),\mathbf{x_{R,i}}(t))  = \|\mathbf{\Delta r_i}\|^{2} = {l_{R}}^2 \\
    &k_{i}(\mathbf{x_{BP}}(t),\mathbf{x_{R,i}}(t),\|\mathbf{T_{i}}\|,\mathbf{u_{i}}) = 0 
    %= \frac{d^2}{{dt}^2}\|\mathbf{\Delta r_i}\|^{2}
\end{aligned}
\label{eq:taut-tether-constraint}
\end{equation}
\noindent where it is assumed for simplicity that the tethers are attached directly to the center of gravity of the robot.

\section{Robot Control}
In this section, we describe our approach for MPC-based closed-loop tracking of desired payload trajectories. We explain the control objectives and formulate the constrained finite horizon optimization problem in order to determine the optimal control moves. The control architecture uitlized for testing in the CAST Arena is highlighted.

\subsection{Objectives}

The system's main functionality is to transport a payload from an initial to a final rest point either on the planet's surface or at height within its atmosphere. The primary control objective is hence to ensure that the payload position $\mathbf{r_{P}}$ reaches a goal position $\mathbf{r_{P,ref}}$ within a given timeframe. By providing time-varying reference positions to the control system,  tracking of predefined waypoints, as could be obtained from a path-planner, can furthermore be achieved.

Unfavourable dynamic modes and disturbances, for instance caused by Martian winds, can however agitate the system and cause oscillations of the payload at rest or during the transport phase. Such oscillations may destabilize the system and decrease the the ability to reliably reach a target position. Therefore, a secondary control reference seeking to minimize the payload swing velocities is set as $\mathbf{\dot{\Theta}_{P,ref}} = [0,0]^\top$.

\subsection{Model Predictive Control Formulation}
An optimal control problem over the prediction horizon $T_p$ is formulated as the following:
\[
\min _{u(\cdot)} \int_{0}^{T_{p}} \Phi_{L}(t, x(t), u(t), d) d t
\]
\textbf{subject to:}
\begin{equation}
\begin{aligned}
x(0) &= x_{0} \\
\dot{x}(t) &=f(t, x(t), u(t), d) \\
g_{i}(t, x(t), u(t), d) &= 0 \\
u_{v} &\geq 0 \\
|u_{R}| &\leq a_u \\
|x(t)| &\leq a_x \\
|y(t)| &\leq a_y \\
\label{eq:mpc}
\end{aligned}    
\end{equation}
\noindent where the states $x(t)=[\mathbf{x_{BP}}^\top,\mathbf{x_{R,1}}^\top ,\hdots, \mathbf{x_{R,n}}^\top]^\top$ and outputs $y(t)=[\mathbf{r_p}^\top, \mathbf{\dot{\Theta}_{P}}^\top]^\top$ for the prediction model assume a total of $n$ UGVs. In the constraint equations above, a disturbance term $d$ accounts for drag model uncertainties and the input vector denoted by $u(t)=[u^\top_{v}(t), u^\top_{R}(t)]^\top$ is composed of inputs to the balloon-payload system in form of the UGV-balloon tether tension forces $u_{v}(t) = [|\mathbf{T_{1}}|, \hdots, |\mathbf{T_{n}}|]^\top$ and inputs to UGVs $u_{R}(t) = [\mathbf{u_{1}}^\top, \hdots, \mathbf{u_{n}}^\top]^\top$.
% Keep this gap please

As the control problem mandates positive tension force magnitudes across the prediction horizon, all virtual inputs $u_{v}(t)$ are bounded to be greater than or equal to zero. Other feasibility constraints are embodied in the lower and upper bounds for the UGV acceleration inputs, state variables such as the robot position and robot velocities and limits on the payload position output variables.

 The constrained optimization problem given above is resolved using MATLAB and Simulink. To reduce the complexity of the MPC state model, we here choose the approach to implement the UGV-tether tension force magnitudes as the additional inputs $u_{v}(t)$. Both inputs $u_{v}(t)$ and $u_{R}(t)$ are optimized simultaneously whilst being subjected to nonlinear equality constraint functions $g_{i}$ in \eqref{eq:mpc}. These constraint functions are chosen as the state dependant constant tether length constraints described by $K_{i}$ as given in \eqref{eq:taut-tether-constraint}, in order to ensure that the physical relation between the robot accelerations and tether forces is maintained.

In \eqref{eq:mpc}, the Lagrange term is chosen to minimize the control input variations $u'(t)$ between time steps and to track the output reference given by $r(t) = [\mathbf{r_{P,ref}}^\top, \mathbf{\dot{\Theta}_{P,ref}}^\top]^\top$. It is given as:
\begin{equation}
\begin{aligned}
\Phi_{L}(t, x(t), u(t), d) =& ||(\mathbf{c_{y}}\odot\mathbf{w_{y}})(r(t)-y(t))||^2 \\ +  &||(\mathbf{c_{u}}\odot \mathbf{w_{\Delta u}})u'(t)||^2
\end{aligned}
\label{eq:lagrange-term}
\end{equation}
\noindent where $\mathbf{w_{y}}$ and $\mathbf{w_{\Delta u}}$ are the optimization weight row vectors for the output tracking and minimal input variation terms, respectively. The row vectors $\mathbf{c_{y}}$ and $\mathbf{c_{u}}$ are used to scale specific outputs and inputs for optimization. The operator $\odot$ denotes an element-wise multiplication with the weight vectors.

%The nonlinear state model is expressed as a function of $x(t)$, UGV acceleration inputs $u_{R}$, virtual inputs $u_{v}$ and a disturbance term $d$ which accounts for drag model uncertainties. 

% for fast compilation using placeholder
\begin{figure}[t]
\begin{center}
\vspace{0.1in}
    \includegraphics[width=\columnwidth]{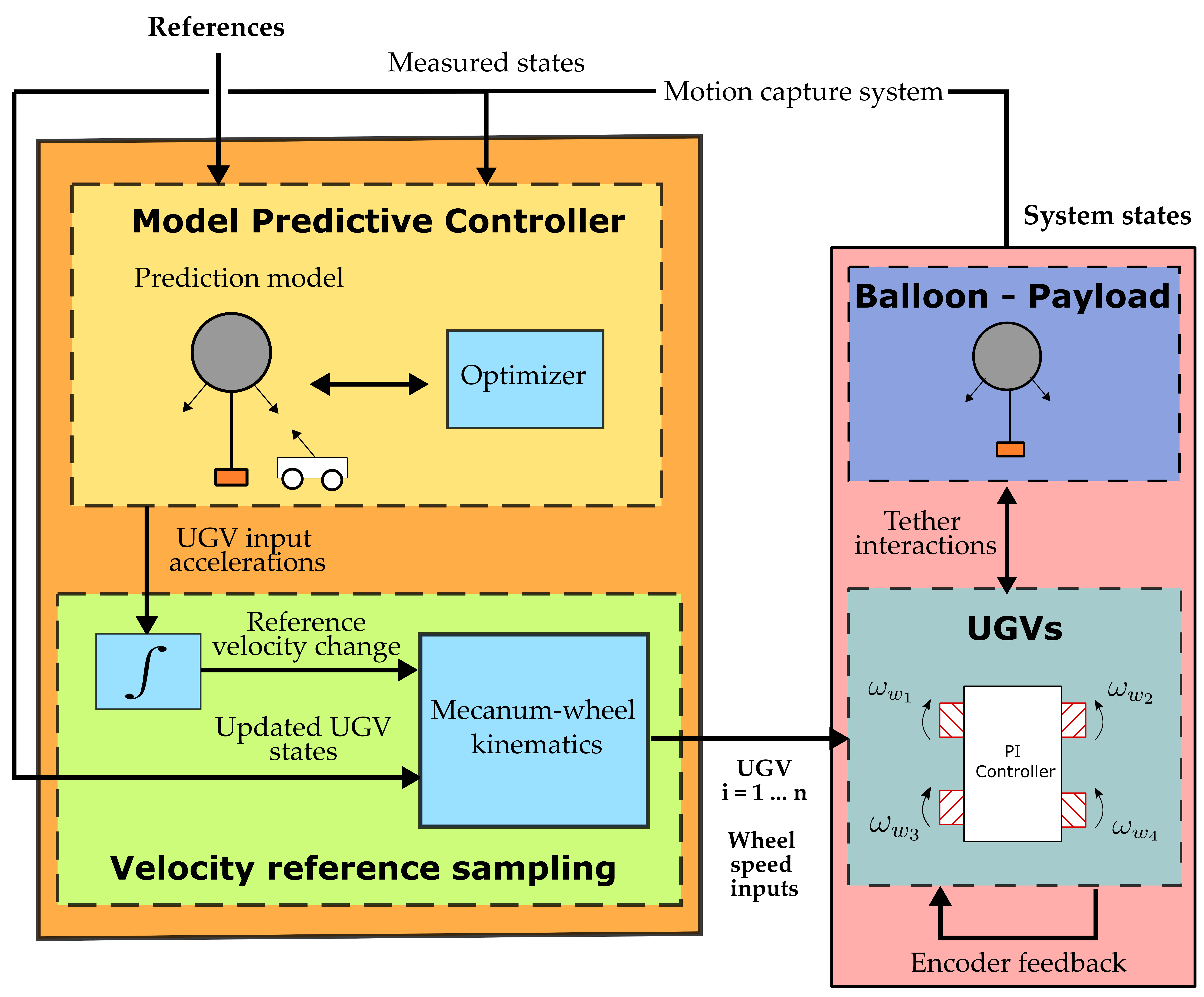}
    \caption{Illustrates an overview of the control architecture.}
    \label{fig:CTRL_arch}
    \vspace{-0.1in}
\end{center}
\end{figure}

\subsection{SDRT-Based Implementation of Control Approach at CAST Arena}
% Due to the large number of possible state trajectories that the system can operate in while tracking the reference values, it is desirable to use optimal control methods to compute the robot control actions. Furthermore, for a scalable system it is beneficial to impose customized optimization criteria which can be adapted for different applications. An added challenge for the control system is that the tethers to the robots must be kept under tension for any robot actions to remain effective, as lightweight strings do not support compression loads. The main controller must in result support combined input and output constraints while minimizing a variable cost-function. Given the previously developed analytical model, it is decided to use Model Predictive Control. \\
The overall control hierarchy is shown in Fig.~\ref{fig:CTRL_arch}. The Model Predictive Controller receives a reference signal from the user or a higher-level trajectory planner. It obtains state measurements which can be complemented by internally computed state estimations if the system states are not fully measurable. For perception and state measurements, the OptiTrack motion capture system is used here. 

The controlled variables are the robot accelerations in the inertial reference frame given by $\mathbf{u_{i}}$, $ ~i \in \left\{ 1,\dots, n\right\}$ as defined prior. These accelerations are piecewise constant over a control interval and are integrated over the controller sample time to provide a velocity change reference signal $\mathbf\Delta {v_{i,ref}}$. In closed loop operation, this reference is summed with the robot initial velocity vector updates $\mathbf{v_{i,0}}$ measured at the beginning of the control interval. Consequently, using the CKM model derived in \eqref{eq:CKM}, a set of four wheel rotational speed references $\mathbf{\omega_{w,ref,i}}$ is obtained. These are sampled at a freqeuency of $\sim$ 20 Hz and transmitted wirelessly from the central control unit to every robot. The robots are driven by an internal PI controller that receives a rotational speed feedback signal from encoders at every wheel and accordingly regulates the motor drive voltage to track the wheel speed references.  

\begin{figure}[t]
\begin{center}
\vspace{0.1in}
    \includegraphics[width=\linewidth]{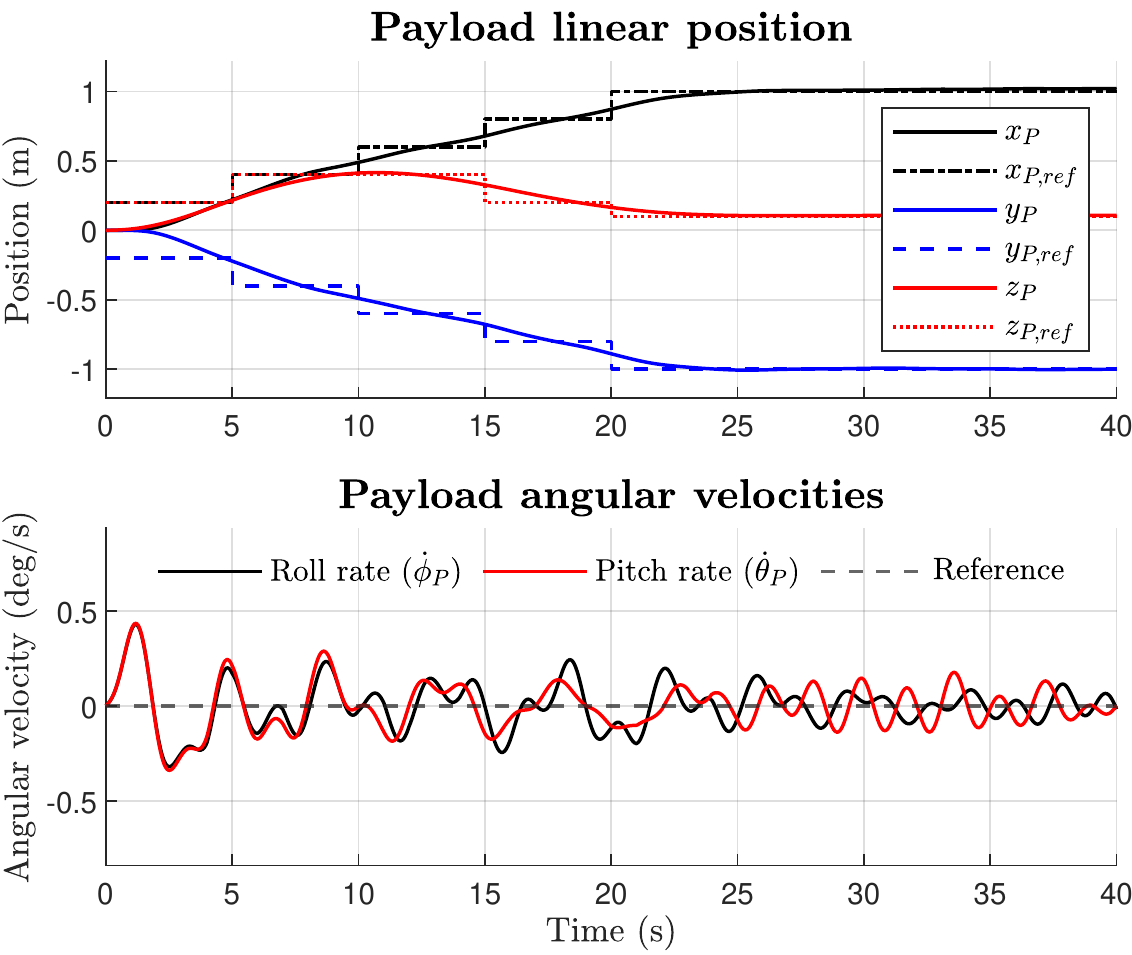}
    \caption{Simulated payload position and tether angular velocities for closed-loop tracking of a time-varying reference trajectory using LTV-MPC.}
    \label{fig:traj_tracking_outputs}
    \vspace{-0.1in}
\end{center}
\end{figure}

\begin{figure}[t]
\begin{center}
\vspace{0.1in}
    \includegraphics[width=\linewidth]{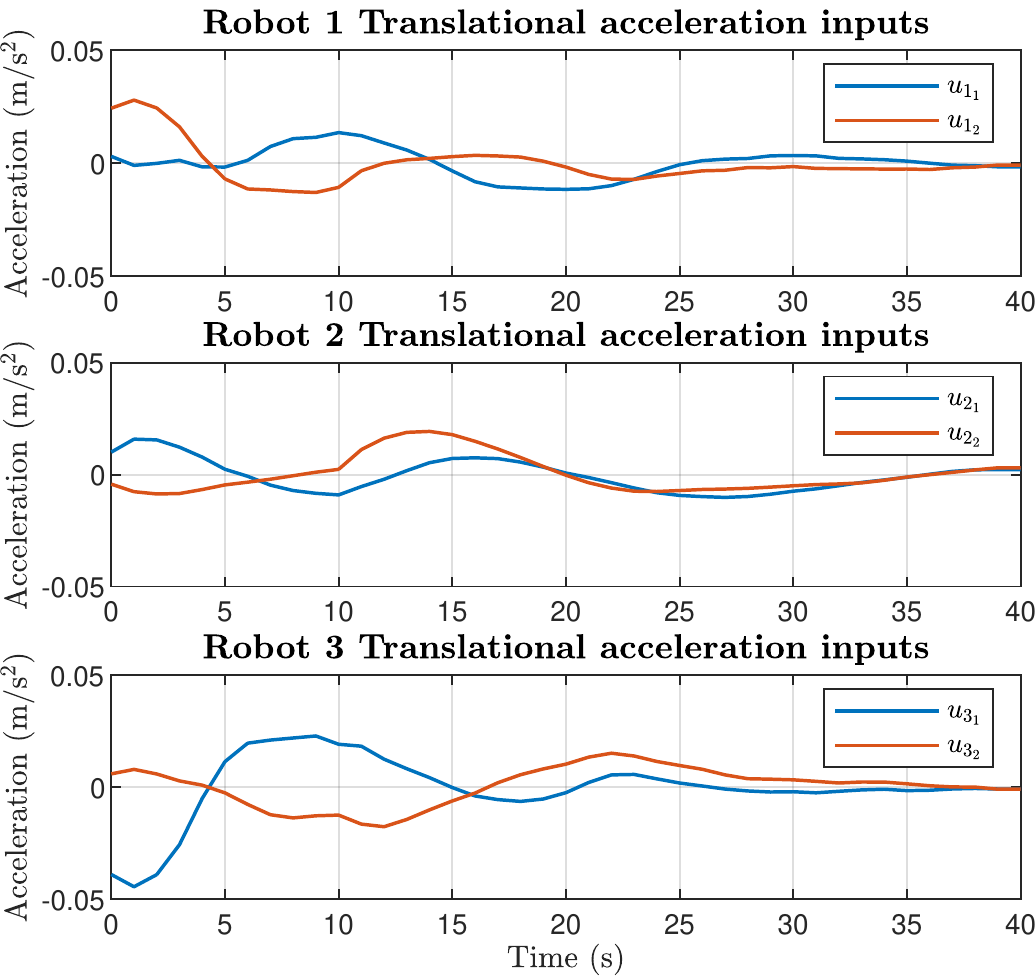}
    \caption{UGV translational acceleration control actions for closed-loop tracking of a time-varying reference trajectory using LTV-MPC}
    \label{fig:traj_tracking_inputs}
    \vspace{-0.1in}
\end{center}
\end{figure}

\section{Results}

In this section, we briefly present our simulation and current experimental results. We show that the closed loop system is able to track time-varying reference waypoints for the payload positions whilst ensuring steady-state stability. First measurements from the physical system are gathered to validate the system model and to allow for controller testing.

\subsection{Simulation}

Closed-loop control simulations for the developed plant model actuated by three ground robots ($n = 3$) are performed. The MPC states, output and constraint functions are derived symbolically in continuous time and are converted to a discrete time formulation.

Due to the slow dynamics of the system, a control sample time on the order of 1.0 seconds is chosen to implement the controller. It was found that the computation time of the nonlinear optimization controller in simulation using sequential quadratic programming solvers used by MATLAB can significantly exceed the control sample time. Thus, a linear-time-varying (LTV) model predictive controller is lastly tested and implemented using the MATLAB MPC Toolbox. By using this control framework, the execution time in simulation is greatly reduced and closed-loop results comparable to the fully nonlinear controller are obtained. 

Equivalent prediction and control horizons of 15 seconds are chosen. The inputs and outputs are scaled over their range of possible values such as to be of unit order for the evaluation of the optimization cost-function defined in \eqref{eq:lagrange-term}. Adding to this, the input variation cost weight vector $\mathbf{w_{\Delta u}}$ is chosen to prioritize small changes in the tether tension force magnitudes in order to reduce the likelihood of temporarily loosing tension in the tethers. The output weight vector $\mathbf{w_{y}}$ is biased towards tracking of the payload swing velocities $\mathbf{\dot{\Theta}_{P}}$ such that steady-state system stability is achieved. 

Throughout our simulations, the absolute translational acceleration inputs $u_{i,1}, u_{i,2}$ are conservatively limited to 0.1 $m/s^2$ to avoid excessive wheel-slip in practice. The UGV rotational acceleration inputs $u_{i,3}$ are not actuated in current experiments and therefore remain zero at every time step.
We implement a time-varying reference signal that imitates a trajectory which could be obtained from a path planner in practical applications. The reference values are updated every five seconds.
In the investigated scenario, the payload is initially at rest on the ground surface with its initial position given as $\mathbf{r_{P,0}} = [0,0,0]^\top$. The payload should then be lifted to and briefly hover at a height of 0.4 m, whilst moving uniformly in the positive longitudinal X and negative longitudinal Y directions. The payload is lastly commanded to continue its longitudinal trajectory whilst moving vertically back down to a height of 0.1 m to reach a steady state position $\mathbf{r_{P,\infty}} = [1,-1,0.1]^\top$ after a total time of 25 seconds. The payload swing angle velocities $\mathbf{\dot{\Theta}_{P,ref}}$ are to be minimized throughout the transport process and after the final position is reached.
To allow for smooth and foresighted tracking of the varying references, the implemented MPC furthermore uses reference previewing, meaning that it accounts for future reference changes when updating the prediction model that is used to determine the optimal control moves at each time step. 

The described reference signal and the obtained closed-loop simulation outputs are given in Fig.~\ref{fig:traj_tracking_outputs}. The corresponding control inputs (UGV accelerations) are given in Fig.~\ref{fig:traj_tracking_inputs}.

%For a step reference and initial position as indicated below, the closed-loop control response of a system initially at rest is given in Figure XXX.

%% \begin{figure}[h]
%%    \begin{center}
%%        \includegraphics[width=\columnwidth]{Images/CONST_REF_FINAL_virt_mass_plot_inputs.pdf}
        %\includegraphics[width=\columnwidth]{example-image-a}
%%        \label{fig:SIM_step-tracking (OPTIONAL)}
%%        \caption{Step response result}
%%    \end{center}
%%    \end{figure}

%%\begin{figure}[h]
%%\begin{center}
%%    \includegraphics[width=\columnwidth]{Images/CONST_REF_FINAL_VIRT_MASS_plot_outputs.pdf}
    %\includegraphics[width=\columnwidth]{example-image-a}
%%    \label{fig:SIM_step-tracking (OPTIONAL)}
%%    \caption{Step response result}
%%\end{center}
%%\end{figure} \\

\begin{figure}[t]
\begin{center}
\vspace{0.1in}
    \includegraphics[width=\columnwidth]{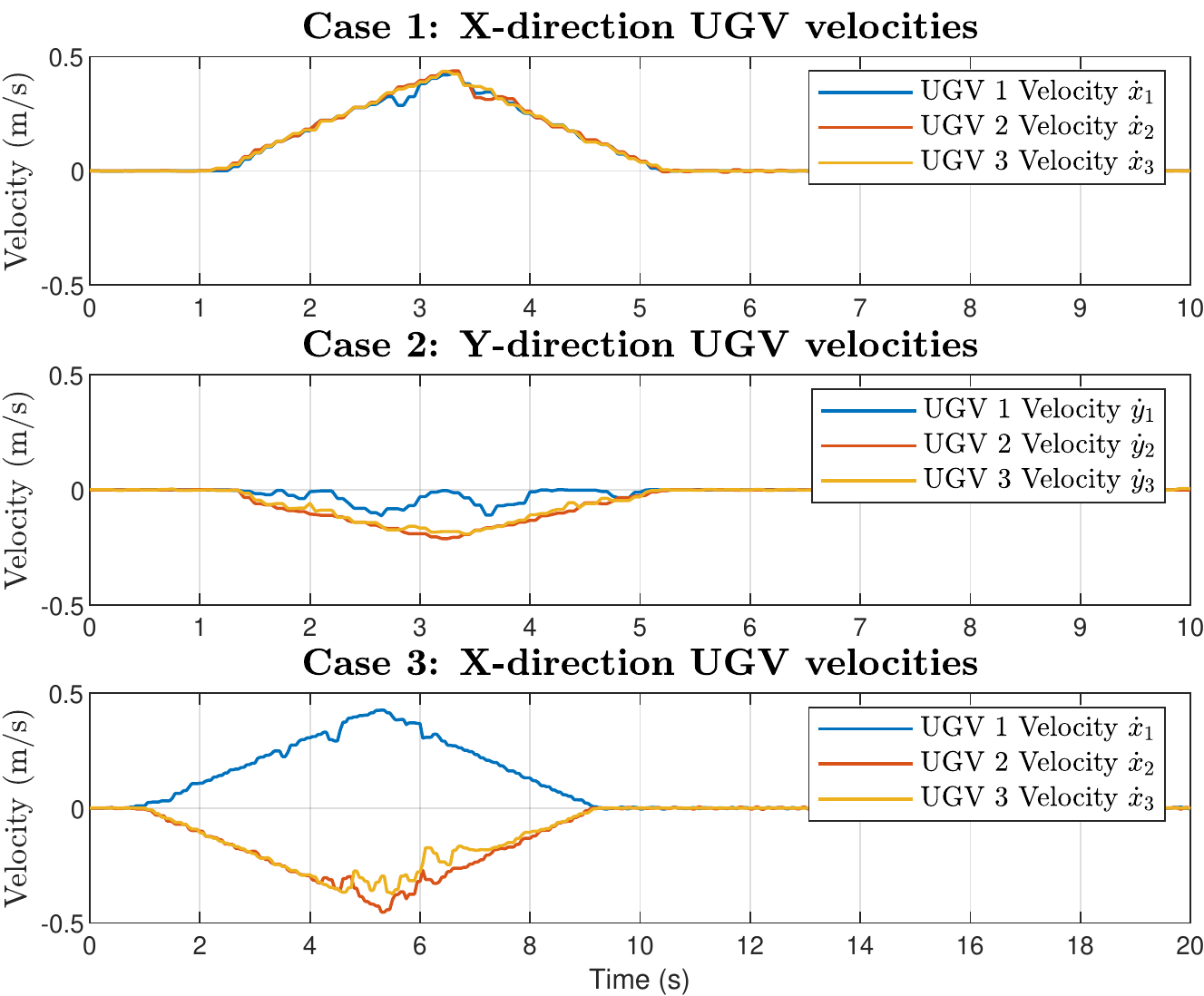}
    \caption{Measured longitudinal UGV velocities for references provided in Case 1: Actuation in longitudinal X direction, Case 2: Actuation in longitudinal Y direction and Case 3: Payload vertical pulling.}
    \label{fig:Val_plot_inputs}
    \vspace{-0.1in}
\end{center}
\end{figure}

\begin{figure}[t]
\begin{center}
\vspace{0.1in}
    \includegraphics[width=\columnwidth]{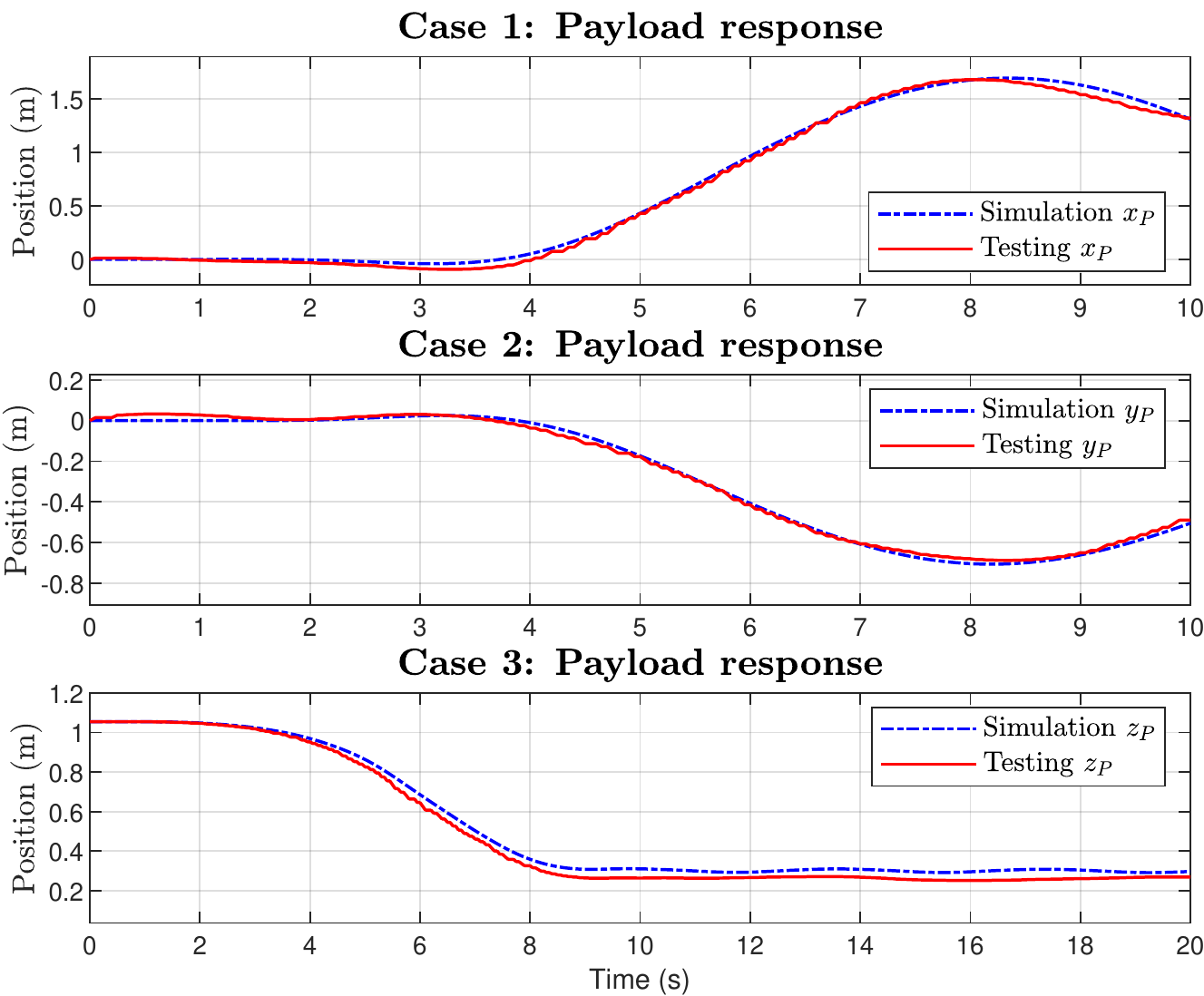}
    \caption{Measured and simulated payload positions in longitudinal X direction, Y-direction and Z-direction for respective experiments Case 1: Actuation in longitudinal X direction, Case 2: Actuation in longitudinal Y direction and Case 3: Payload vertical pulling.}
    \label{fig:Val_plot_outputs}
    \vspace{-0.1in}
\end{center}
\end{figure}

\subsection{Experiments}

\begin{figure}
    \centering
    \vspace{0.1in}
    \includegraphics[width = \linewidth]{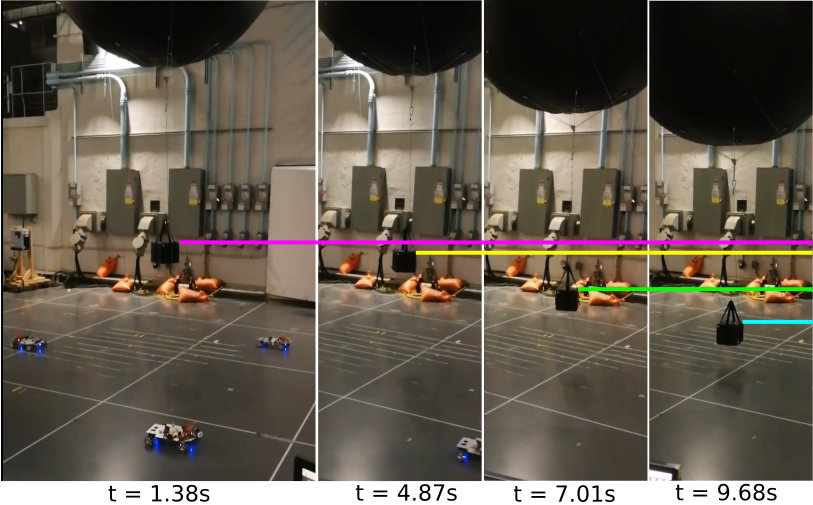}
    \caption{Snapshots of the video taken during the experiment. The lines indicate the upper side of the payload, showing how the payload elevation can be adjusted down (Test Case 3) by moving the ground robots outwards relative to the payload position.}
    \vspace{-0.1in}
    \label{fig:baloon_snapshots}
\end{figure}

Experiments are performed where state responses of the prototype to different feed-forward control inputs are measured and compared to simulation results. The three rovers are commanded with predefined velocity profiles constructed from acceleration and deceleration inputs of piece-wise constant and equal magnitude. We consider three experiment cases: In Case 1, the UGVs are actuated equally only in the longitudinal X-direction, whereas in Case 2, all rovers are equally actuated only in the longitudinal Y-direction. In Case 3, the payload is to be moved down vertically by moving the rovers apart longitudinally, that is by providing an X-direction velocity reference to UGV 1 that is equal in magnitude but opposite in direction to that of UGVs 2 and 3. The snapshots of the Test Case 3 can be seen in Fig.~\ref{fig:baloon_snapshots}, showing that the payload was moved down vertically.

The realized robot velocities in the inertial plane and all remaining system states and outputs are measured using OptiTrack. Filtered UGV acceleration profiles are then derived from the velocity measurements and given as a time-varying input to the simulation model, where the simulated plant state and output trajectories are extracted. Figure \ref{fig:Val_plot_inputs} shows measured robot velocity profiles in the actuated directions for the three different testing cases and Fig.~\ref{fig:Val_plot_outputs} compares the corresponding simulated and measured payload outputs. 

It is observed that due to wheel slip disturbances, the measured velocity profiles in the Y-direction shown in Fig.~\ref{fig:Val_plot_inputs} for Case 2  are not equal for all robots. Using the validation approach explained above, this behavior is however mirrored in the UGV acceleration inputs provided to the simulation framework. It is overall seen that good agreement between the tested and simulated outputs exists for these three decoupled cases. Further open-loop and closed-loop control tests are ongoing.

\section{CONCLUSIONS}

In this work, we covered an actuation and control design framework for steering a balloon that is rope-driven by a number of ground vehicles in martian craters. Our approach is based on the coordinated servoing of these unmanned ground vehicles to regulate tension forces in the cable-driven balloon. The loitering and trajectory tracking of an underactuated payload suspended from the balloon was considered as the overarching objective of the paper which was achieved using a model predictive control design.

While the feasibility of performing airborne and ground manipulation, perception, and reconnaissance using wheeled rovers or unmanned aerial vehicles have been evaluated before, these robots can face major challenges operating in Mars craters. In our future works, we plan to further validate the control robustness in experiments and to improve our balloon model to accommodate uncertainties led by environmental effects on the balloon. Balloon-based solutions possess merits that make them extremely attractive, e.g., simple operation mechanism and endured operation time. However, many hurdles remain to overcome to achieve robust loitering and tracking performance by balloons in Mars craters.

\addtolength{\textheight}{-12cm}   % This command serves to balance the column lengths
                                  % on the last page of the document manually. It shortens
                                  % the textheight of the last page by a suitable amount.
                                  % This command does not take effect until the next page
                                  % so it should come on the page before the last. Make
                                  % sure that you do not shorten the textheight too much.

%%%%%%%%%%%%%%%%%%%%%%%%%%%%%%%%%%%%%%%%%%%%%%%%%%%%%%%%%%%%%%%%%%%%%%%%%%%%%%%%

\bibliographystyle{IEEE}
\bibliography{bibliography,references}

\end{document}